\documentclass[journal]{IEEEtran}
\usepackage[utf8]{inputenc}
\usepackage[T1]{fontenc}
\usepackage{siunitx}
\usepackage[colorlinks=true, allcolors=blue]{hyperref}
\usepackage{amsmath}
\usepackage{amsfonts}
\usepackage{soul} 
\usepackage[caption=false]{subfig}
\usepackage{stfloats}
\usepackage{float} 

\usepackage{tikz,xcolor}
\definecolor{lime}{HTML}{A6CE39}
\DeclareRobustCommand{\orcidicon}
{
    \begin{tikzpicture}
    \draw[lime, fill=lime] (0,0) circle [radius=0.16] 
    node[white] {{\fontfamily{qag}\selectfont \tiny ID}};    \draw[white, fill=white] (-0.0625,0.095) circle [radius=0.007];    
    \end{tikzpicture}
    \hspace{0mm}}
\foreach \x in {A, ..., Z}{%
    \expandafter\xdef\csname orcid\x\endcsname{\noexpand\href{https://orcid.org/\csname orcidauthor\x\endcsname}{\noexpand\orcidicon}}
}

\begin{document}

\title{Traffic Flow Simulation for Autonomous Driving}

\author{Junfeng Li, Changqing Yan}
\maketitle
\begin{abstract}
A traffic system is a random and complex large system, which is difficult to conduct repeated modelling and control research in a real traffic environment. With the development of automatic driving technology, the requirements for testing and evaluating the development of automatic driving technology are getting higher and higher, so the application of computer technology for traffic simulation has become a very effective technical means\cite{young2020unreal}.

Based on the micro-traffic flow modelling, this paper adopts the vehicle motion model based on cellular automata and the theory of bicycle intelligence to build the simulation environment of autonomous vehicle flow. The architecture of autonomous vehicles is generally divided into a perception system, decision system and control system. The perception system is generally divided into many subsystems, responsible for autonomous vehicle positioning, obstacle recognition, traffic signal detection and recognition and other tasks. Decision systems are typically divided into many subsystems that are responsible for tasks such as path planning, path planning, behavior selection, motion planning, and control. The control system is the basis of the self-driving car, and each control system of the vehicle needs to be connected with the decision-making system through the bus, and can accurately control the acceleration degree, braking degree, steering amplitude, lighting control and other driving actions according to the bus instructions issued by the decision-making system, so as to achieve the autonomous driving of the vehicle\cite{dudzik2022autonomous}.
\end{abstract}


\IEEEpeerreviewmaketitle

\section{Introduction}
With the development of autonomous vehicles (AV), safety evaluation of control strategy becomes more and more important. Testing with actual vehicles is dangerous, and hi-fi simulators,
such as Airsim, will take a lot of time. Therefore, a safe, effective and low-cost evaluation method is urgently needed in practice. Game engines are increasingly being used by the autonomous vehicle community as a simulation platform to develop vehicle control systems and test environments \cite{chance2022determinism}.

This paper introduces a micro-traffic model based on UE4-generated traffic flow, in which each vehicle is regarded as an independent entity and path planning is carried out by a global scheduler, to simulate real traffic scenes and conditions. 
This paper implemented the traffic flow based on the spline method in UE4 and the grid method in AI and evaluated the results of the two methods, which are helpful to optimize the traffic flow generation.

\section{Research Targets}
\label{Research Targets}
\subsection{Traffic flow definition}
Traffic flow can be thought of as a flow: vehicles in the flow share similar goals and rules of behaviour, interacting with their neighbours while maintaining their respective driving characteristics \cite{wang1995enhanced}.
\subsection{Target}
A reasonable standard for generating traffic flow:
(Each vehicle is treated as a separate agent.)
\begin{itemize}
\item The vehicle is on the right road and will not go off the road.
\item The vehicle will avoid obstacles and will not collide with roadblocks and other vehicles.
\item The vehicle will accurately park at the destination (parking space) and will not deviate from the endpoint. 
\end{itemize}

\section{Traffic flow simulation modelling}
\label{sec:methodology}
Simulation is crucial for the application of autonomous driving in the real world. 
\subsection{Macro Traffic Simulator}
The macro approach is also called the continuum approach. Describe vehicle behaviour and interactions in low detail: traffic flow is represented by a continuum of speed, flow, density, etc. The macro approach is primarily a vehicle simulation on a large-scale road network designed for efficiency, with an emphasis on reproducing aggregate behaviour measured collectively in terms of flow density and traffic flow.
\subsection{Microscopic Traffic Simulator}
Microsimulation generates vehicle motion at a high level of detail: each vehicle is treated as a discrete agent that satisfies certain control rules \cite{yang1996microscopic}.

Early microscopic models include the cellular automata model and  vehicle following model. 
In this project, the cellular automata model is used to model traffic flow. The motion of the vehicle in the cellular automata model is described by the evolution rules in the pre-specified time, space and state variables \cite{levin2016cell}. 
Specifically, the road is discretized into units, and the model determines when the vehicle moves from the current unit to the next. Due to its simplicity, the cellular automata model has high computational efficiency and can simulate a large number of vehicles on a large road network. However, due to its discrete nature, the generated virtual car can only reproduce a limited number of real car behaviors.

The description of traffic flow in the microscopic vehicle flow model is that the microscopic behaviors of vehicles with a single vehicle as the basic unit on the road, such as following, overtaking and lane change, can be simulated. Therefore, such models are also called behaviour-based models \cite{nv2020study}
.

\section{Autonomous Driving}
\label{sec:setup}

\subsection{planning demand}
Planning can be divided into Routing, Behavioral Decision, and Motion Planning.

Routing: It is a global path planning, which can be simply understood as traditional map navigation + high-definition map (including lane information and traffic rules, etc.);

Behavioural decision: decide whether the vehicle follows, waits and avoids when encountering traffic lights and pedestrians, and interacts with other vehicles at intersections;

Motion Planning: It is local path planning, which is the expected path of self-driving cars in the future, and needs to meet the requirements of automotive kinematics, dynamics, comfort and collision free.

The task of trajectory planning is to calculate a collision-free trajectory (including path and speed information) to ensure that the vehicle is driven safely from the starting point to the destination and as efficiently as possible \cite{yang2019efficient}.

optimization objectives include:

Safety: Avoid collisions with obstacles in the scene; For dynamic obstacles, reduce their future collision risk due to the uncertainty of their future movement \cite{chae2018virtual}; 

Stability: Due to the large inertia and poor flexibility of the vehicle, the expected trajectory needs to ensure the physical feasibility of the vehicle and the stability of the controller;

Comfort: Considering the comfort of the occupants, it is necessary to ensure the driving comfort of the vehicle while meeting safety and stability, including acceleration and deceleration and steering;

Timeliness: While meeting safety and stability, it ensures that the vehicle drives at a faster speed, to reach the destination in a shorter time.

In a real-world scenario, the planning process needs to consider various physical constraints, including but not limited to:

Acceleration and deceleration constraints: limited by the performance limits of the power system and braking system, and the safety and comfort of the driver;

Non-integrity constraints: the vehicle has three degrees of freedom of motion, but only two degrees of control freedom, and its non-integrity constraints determine the physical feasibility of the trajectory;

Dynamic constraints: considering the dynamic characteristics and body stability of the vehicle, the curvature and yaw angular velocity during driving have certain constraints;

\subsection{AIPerception}

This is Carla's most commonly used vehicle tool for perceiving its surroundings \cite{liu2002algorithm};

\subsubsection{how to use it?}
\hspace*{\fill}

1. Hang the perception component AIPerceptionComponent to the Controller of the AI agent;

2. Select the type of perception ability (vision, hearing, touch, injury perception, etc.) that the AI needs to have on the perception component and set the relevant attribute parameters;

3. When the target object is perceived, the perception component throws the event AIPerceptionComponent::OnPerceptionUpdated, which tells which target object Actors are perceived.

\subsubsection{System components}
\hspace*{\fill}

Step 1: Initialize the perception component, call AIPerceptionComponent:: OnRegister(), traverse all perception configurations SensesConfig in the component, and register all AISensions in AIPerceptionSystem; At the same time, add itself to the ListenerContainer of the AIPerceptionSystem.

Step 2: AIPerceptionSystem collects the information transmitted during the initialization of AIPerceptionComponent. After all sensing components are initialized, AIPerceptionSystem knows all listeners with sensing behavior in the current world, ListenerContainers, and all required sensing capabilities Senses.

Step 3: AIPerceptionSystem uniformly processes all sensing capabilities and listeners in its Tick method, traversing Senses and ListenerContainer every frame, updating ListenerContainer location information, and executing the Tick method for each Sense.

Step 4: In the Update method of the Sense subclass, execute specific perceptual ability judgment logic, generally traversing all mappings between listeners and stimulus sources, such as AISightQuery for visual perception and ListenersMap for auditory perception. When a pair of mapped stimulus sources can be met (perceived) by the listener's perception conditions, register the modified stimulus source and its stimuli in the StimulusToProcess of the modified listener for processing by the listener.

Step 5: AIPerceptionSystem calls the ProcessStimuli method every frame, traverses all newly registered StimuliToProcess stimuli in the previous frame, and updates the relevant information of all stimulus source agents, PerceptialData; Identify all stimuli that need to be updated (newly perceived or removed) and throw the event PerceptionUpdatedDelegate.

At the same time, the ActorsPerceptionUpdated() method of AIController will be called, with the parameter being the stimulus source Actor to be updated, notifying the AI perception system that a certain stimulus source Actor has been updated.

After processing all newly registered stimuli in the previous frame, clear StimuliToProcess and start the registration collection of perceptual stimuli in the next frame.

\subsection{Route Planning}

The route planning subsystem is responsible for calculating the route through the road network from the initial position of the self driving vehicle to the final position defined by the user operator. If a weighted directed graph is used to represent the road network, and its edge weights represent the cost of passing through a road segment, then the problem of calculating a route can be reduced to finding the shortest path in the weighted directed graph \cite{bhattarai2020deep}
.

The path planning methods in road networks provide different trade-offs in terms of query time, preprocessing time, space utilization, and robustness to input changes. They can be mainly divided into four categories: goal-directed, separator based, hierarchical, bounded top, and combinations.

\subsubsection{Goal-Directed Techniques}

Target oriented technology guides the search from the source vertex to the target vertex by avoiding scanning vertices that are not in the direction of the target vertex \cite{rao2020research}. A * is a classic goal oriented shortest path algorithm. Compared to the Dijkstra algorithm, this algorithm uses a lower distance function on each vertex, allowing vertices closer to the target to be scanned earlier and achieving better performance. The ALT (A *, landmark, and triangle inequality) algorithm enhances A * by selecting a small group of vertices as landmarks. In the preprocessing stage, calculate the distance between all landmarks and all vertices. During the query phase, use triangle inequalities containing landmarks to estimate the effective lower bound distance of any vertex. Query performance and correctness depend on whether vertices are wisely selected as markers. Another target orientation algorithm is Arc Flags. In the preprocessing stage, the graph is divided into units with a small number of boundary vertices and balanced (i.e. similar) vertices. By growing the Shortest-path tree backward from each boundary vertex, set the ith flag for all arcs (or edges) of the tree, and calculate the arc flag of unit i. In the query phase, the algorithm will trim edges that do not have flags set for cells containing target vertices. The arc flags method has a high preprocessing time, but the query time is the fastest among target oriented techniques.

\subsubsection{Separator-Based Techniques}
Target oriented technology guides the search from the source vertex to the target vertex by avoiding scanning vertices that are not in the direction of the target vertex. A * is a classic goal oriented shortest path algorithm. Compared to the Dijkstra algorithm, this algorithm uses a lower distance function on each vertex, allowing vertices closer to the target to be scanned earlier and achieving better performance. The ALT (A *, landmark, and triangle inequality) algorithm enhances A * by selecting a small group of vertices as landmarks. In the preprocessing stage, calculate the distance between all landmarks and all vertices. During the query phase, use triangle inequalities containing landmarks to estimate the effective lower bound distance of any vertex. Query performance and correctness depend on whether vertices are wisely selected as markers. Another target orientation algorithm is Arc Flags. In the preprocessing stage, the graph is divided into units with a small number of boundary vertices and balanced (i.e. similar) vertices. By growing the Shortest-path tree backward from each boundary vertex, set the ith flag for all arcs (or edges) of the tree, and calculate the arc flag of unit i. In the query phase, the algorithm will trim edges that do not have flags set for cells containing target vertices. The arc flags method has a high preprocessing time, but the query time is the fastest among target oriented techniques.
\subsubsection{Hierarchical Techniques}
Hierarchy technology utilizes the inherent hierarchical structure of road networks, where major roads (such as highways) form a small backbone subnet. Once the distance between the source and target vertices is significant, the query algorithm only scans the vertices of the subnet. The preprocessing stage calculates the importance of vertices or edges based on the actual shortest path structure. The CH (compression hierarchies) algorithm is a layered technique that implements the idea of creating shortcuts to skip vertices of lower importance. It repeatedly performs vertex compression operations. If the shortest path in the graph is unique and contains the vertices to be compressed, the least important vertices are removed from the graph and shortcuts are created between each pair of adjacent vertices. CH is universal and can serve as a building block for other point-to-point algorithms and extended queries. REACH algorithm is a hierarchical technique, which calculates vertex centrality measures (REACH values) in the preprocessing phase, and uses them to trim Dijkstra based Bidirectional search in the query phase. Let P be the shortest path from the source vertex s to the target vertex t containing vertex v. The arrival value of v relative to P is r (v, P)=min {distance (s, v), distance (v, t)}.
\subsubsection{Bounded-Hop Techniques}
Bounded hop technology pre-calculates the distance between vertex pairs by adding virtual shortcuts to the graph. 
Due to the infeasibility of pre-calculating the distance between all vertex pairs for large networks, the goal of bounded hop technology is to obtain the length of any virtual path with very few hops.

\subsection{Motion Planning}

The Motion planning subsystem is responsible for calculating the path or track from the current state of the autonomous vehicle to the next local target state defined by the behavior selection subsystem. The motion scheme implements local driving behavior, meets the Kinematics and dynamics constraints of the car, provides comfort for passengers, and avoids collision with static and moving obstacles in the environment.

A motion plan can be a path or trajectory. A path is a sequence of car states that does not define how car states evolve over time. This task can be delegated to other subsystems (such as behavior selection subsystems) or the velocity profile can be defined as a function of curvature and proximity to obstacles. A trajectory is a path that specifies the evolution of a car's state over time.
\subsection{Global and Local Planning}
Global: Enter a destination and the algorithm will automatically calculate the optimal global route

Local: Focus only on routes over a period of time

Simulating real-world scenarios requires a combination of these two planning methods \cite{leudet2019ailivesim}.
Global planning belongs to static planning, and local planning belongs to Dynamic programming. Global path planning requires mastering all environmental information and planning paths based on all information from the environmental map; Local path planning only requires sensors to collect real-time environmental information, understand environmental map information, and then determine the location of the map and the distribution of local obstacles, to select the optimal path from the current node to a certain sub-target node.

\subsection{Common used planning algorithms}
\subsubsection{Artificial potential field method}

\hspace*{\fill}

It is assumed that the driving target point will generate gravity on the vehicle, while the obstacles will generate repulsion on the vehicle, and the vehicle will be controlled to move along the "potential valley" between the "potential peaks" in the potential field. Among them, gravity is proportional to the distance from the vehicle to the moving target point, and repulsion is inversely proportional to the distance from the vehicle to the obstacle. The driving speed and direction of the vehicle are controlled by solving the resultant force of gravity and repulsion as the external force of the vehicle. This method has the advantages of an easy mathematical expression, fast reaction speed, and easy implementation of closed-loop control between the algorithm and environment. However, it is prone to locally optimal solutions during the solving process, leading to deadlock phenomenon

\subsubsection{Motion planning method based on graph search}
\hspace*{\fill}

The initial pose and target pose of the vehicle are mapped to a state space, and then the state space is Discretization to form a graph, from which the optimal trajectory satisfying the constraint conditions is searched. At present, the mainstream methods mainly include the Voronoi map, grid map and cost map, Lattice state map, driving channel map, etc \cite{han2021new}. In order to balance real-time performance and obstacle constraint space processing ability, Lattice and channel graph methods are generally used to generate safe trajectories.

\subsubsection{A Path Planning Algorithm Based on Random Sampling}
\hspace*{\fill}

The basic idea of the random sampling method is to randomly sample in the configuration space and screen out the optimal sampling points that meet the performance requirements. It has probability Completeness, but its biggest disadvantage is poor comfort, and the calculation efficiency decreases with the increase of the number of obstacles

\subsubsection{Based on curve interpolation method}
\hspace*{\fill}

Based on the pre-constructed curve type, according to the expected state of the vehicle (position, speed, acceleration, heading angle, etc.), these expected values are substituted into the curve type as boundary conditions for Equation solving to obtain the correlation coefficient of the curve. After all coefficients are calculated, the curve trajectory planning is completed

\section{Implementation method}
\label{sec:results}

\begin{itemize}
\item Preliminary design:
\end{itemize}
Using animated blueprints, the motion of the car can be realized, including the rotation of the tires.

The plan is to use Inverse kinematics, where IK redirection animations allow tires to move without disassociating from the body: As opposed to FK, the tire is treated as a skeletal plug-in and robot kinematics equations are used to determine joint parameters and move the tire to desired positions all over the world.

The team uses UE4 as the simulation environment. Based on UE4, there are the following two methods to realize bicycle intelligence. The team has explored and evaluated both methods, and the experimental results are as follows:
\subsection{Automatic pathfinding based on Spline}
\subsubsection{Calculate steering} 
The core of Spline’s autonomous driving is the real-time computing that updates the Steering of
the car.
\begin{itemize}
\item First get the position of the current car, namely the geometric centre of the car.
\item Find out the nearest distance from the car position to the curve by the algorithm, find the tangent line of the point on the curve, make normalization processing, and multiply by the smoothing coefficient (the smaller the coefficient, the higher the sensitivity of the car).
\item Translate the vector to the position of the car to get the closest point from the end point of the vector to the curve and get the vector pointing to the change point, namely, the rotation Angle of the car. Use the clamp function to convert this Angle (-90,90°) into the steering Angle (-1,1).
\item Depending on collision detection, when the car collides with the target actor (such as the wall of the parking point, ground lock, etc.), the car will stop.
\end{itemize}

\begin{figure}[H]
\centering
\includegraphics[width=0.5\textwidth]{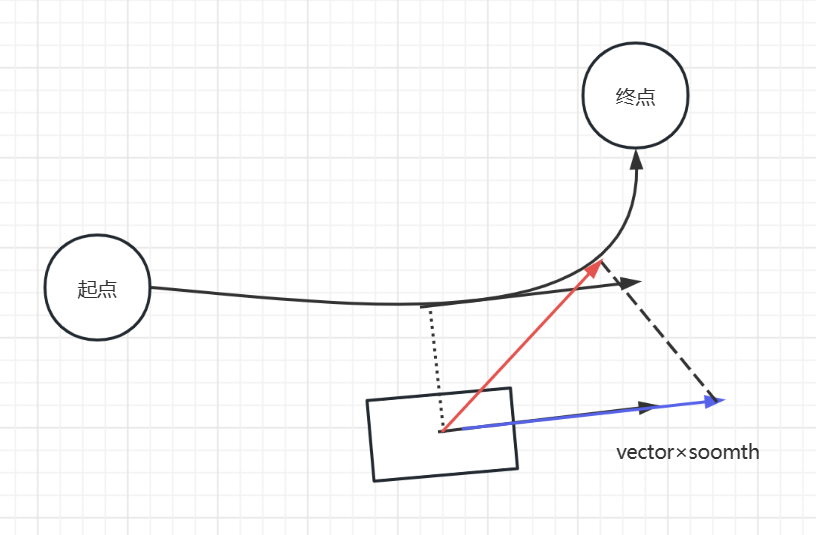}
\subfloat{\label{fig:P1}Fig.1  Spline}
\end{figure}

In this environment, the car can drive on the road automatically, avoid obstacles and cars in front of it, and realize automatic operations such as turning, going straight and parking.

\begin{figure}[H]
\centering
\includegraphics[width=0.5\textwidth]{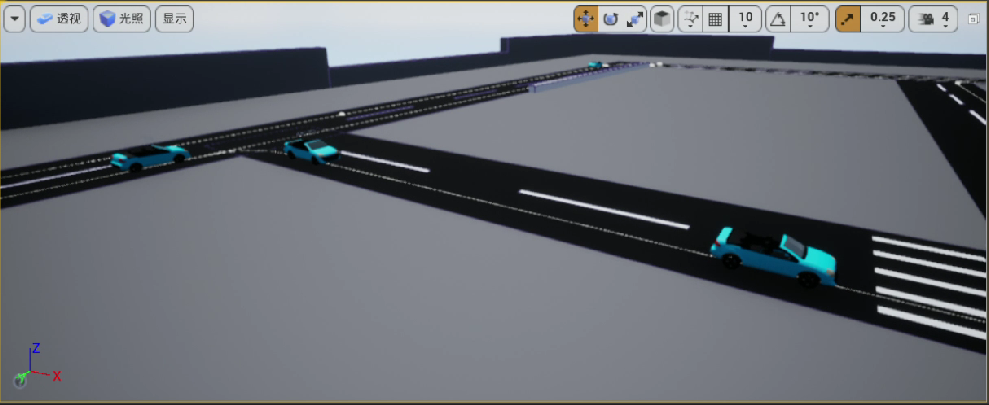}
\subfloat{\label{fig:P2}Fig.2  Spline Implement}
\end{figure}
\subsection{Automatic pathfinding based on grid}

Depending on AIMoveTo: This method automatically generates an actor’s arrival Path when given
a starting point and an ending point (Path Start and Path End). Plan routes automatically using the AIMoveTo method that comes with UE4, and plan car routes using a grid.
\begin{figure}[H]
\centering
\includegraphics[width=0.5\textwidth]{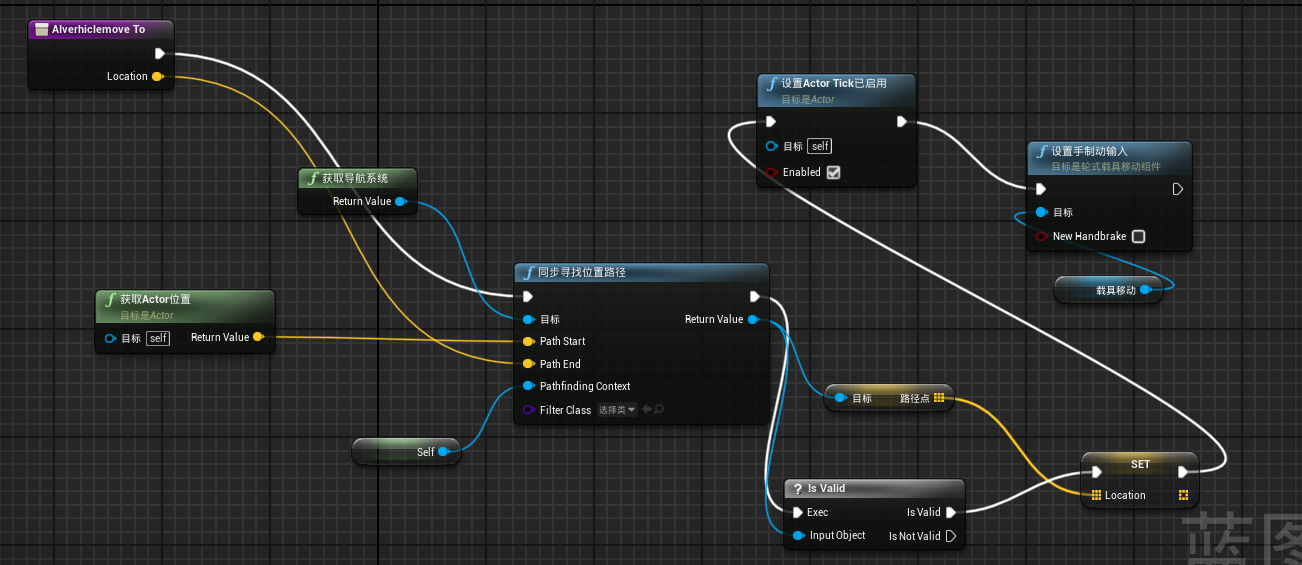}
\subfloat{\label{fig:P3}Fig.3  Blueprint}
\end{figure}
The current effect is to enable the vehicle to automatically avoid some simple obstacles and automatically navigate to a specified location. Limit vehicle travel distance according to grid navigation: the grey part is the part that AI recognizes cannot pass, while the green part is the part that the vehicle can pass. The effective design of the navigation grid can avoid many obstacles, such as avoiding the situation where the car hits the edge of a slope instead of driving along it.
\begin{figure}[H]
\centering
\includegraphics[width=0.5\textwidth]{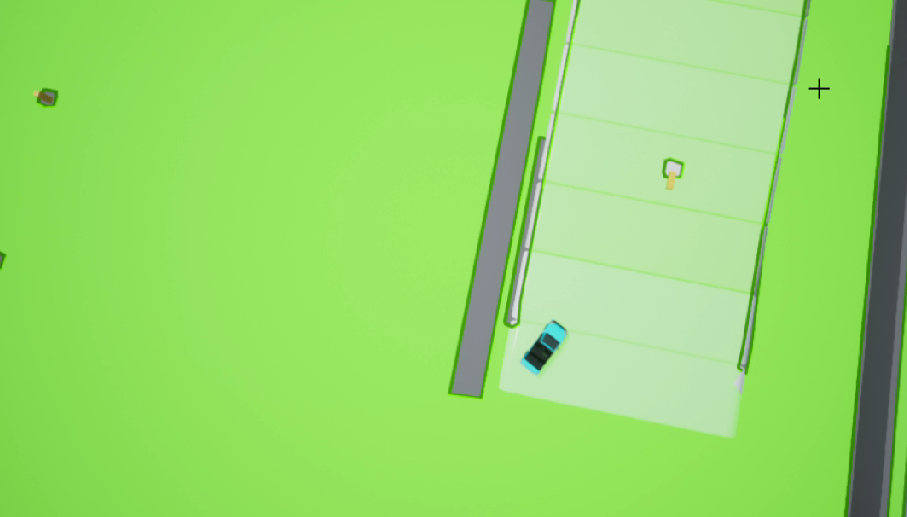}
\subfloat{\label{fig:P4}Fig.4  Implement}
\end{figure}
\section{Evaluation and conclusion}
\label{sec:discussion}
\subsection{Driving depend on Spline}

Result: Generate a traffic flow based on the spline, and cycle along the planned spline, which can detect the traffic light and judge whether to move forward. And perform turning, parking and other operations.

Advantages: The route planning is stable, the vehicle will not deviate from the track, the path is already known in advance, and the running speed will be faster.

Drawback: If the map is updated before the Spline curve is manually added, the car cannot realize autonomous driving, which relies A lot on the spline. 
\subsection{Driving depends on grid}

Result: The AI method is used to make decisions, the car driving route is planned using the grid body, and the vehicle moving distance is restricted according to the grid body navigation: the grey part is the part that the AI can not pass, and the green part is the part that the vehicle can pass.

Advantages: there is no need to plan the path, and the path can be automatically generated by the algorithm, which has little dependence on the simulated map, and has good flexibility, and the algorithm can be improved and the optimization space is large.
Through routing algorithms such as A*, the car itself can have obstacle avoidance and wayfinding functions, and automatically detect the front actor to achieve obstacle avoidance. Putting the car into any environment can achieve autonomous driving.

Drawback: The logic of the vehicle running requires the control of steering and throttle switches. And if there is an obstacle in front of the car or the navigation grid body is forbidden to move in the area, the throttle needs to be closed. (or reduce the number of actions according to the number of frames), which facilitates turning and alignment, and when the car is about to reach the specified position, it needs to slow down in advance to ensure that it stops at the specified position and does not pass it.

In the future, we will apply various AI technologies, such as knowledge graphs for semantic reasoning, classification, and computer vision for objects detection to improve this work.

\newpage
\bibliographystyle{IEEEtran}
\bibliography{references}

\begin{thebibliography}{10}
\providecommand{\url}[1]{#1}
\csname url@samestyle\endcsname
\providecommand{\newblock}{\relax}
\providecommand{\bibinfo}[2]{#2}
\providecommand{\BIBentrySTDinterwordspacing}{\spaceskip=0pt\relax}
\providecommand{\BIBentryALTinterwordstretchfactor}{4}
\providecommand{\BIBentryALTinterwordspacing}{\spaceskip=\fontdimen2\font plus
\BIBentryALTinterwordstretchfactor\fontdimen3\font minus
  \fontdimen4\font\relax}
\providecommand{\BIBforeignlanguage}[2]{{%
\expandafter\ifx\csname l@#1\endcsname\relax
\typeout{** WARNING: IEEEtran.bst: No hyphenation pattern has been}%
\typeout{** loaded for the language `#1'. Using the pattern for}%
\typeout{** the default language instead.}%
\else
\language=\csname l@#1\endcsname
\fi
#2}}
\providecommand{\BIBdecl}{\relax}
\BIBdecl

\bibitem{young2020unreal}
P.~Young, S.~Kysar, and J.~P. Bos, ``Unreal as a simulation environment for
  off-road autonomy,'' in \emph{Autonomous Systems: Sensors, Processing, and
  Security for Vehicles and Infrastructure 2020}, vol. 11415.\hskip 1em plus
  0.5em minus 0.4em\relax SPIE, 2020, pp. 113--120.

\bibitem{dudzik2022autonomous}
M.~C. Dudzik, S.~M. Jameson, and T.~J. Axenson, ``Autonomous systems: Sensors,
  processing and security for ground, air, sea and space vehicles and
  infrastructure 2022,'' in \emph{Proc. of SPIE Vol}, vol. 12115, 2022, pp.
  1\,211\,501--1.

\bibitem{chance2022determinism}
G.~Chance, A.~Ghobrial, K.~McAreavey, S.~Lemaignan, T.~Pipe, and K.~Eder, ``On
  determinism of game engines used for simulation-based autonomous vehicle
  verification,'' \emph{IEEE Transactions on Intelligent Transportation
  Systems}, 2022.

\bibitem{wang1995enhanced}
P.~T. Wang and R.~A. Glassco, ``Enhanced thoreau traffic simulation for
  intelligent transportation systems (its),'' in \emph{Proceedings of the 27th
  conference on Winter simulation}, 1995, pp. 1110--1115.

\bibitem{yang1996microscopic}
Q.~I. Yang and H.~N. Koutsopoulos, ``A microscopic traffic simulator for
  evaluation of dynamic traffic management systems,'' \emph{Transportation
  Research Part C: Emerging Technologies}, vol.~4, no.~3, pp. 113--129, 1996.

\bibitem{levin2016cell}
M.~W. Levin and S.~D. Boyles, ``A cell transmission model for dynamic lane
  reversal with autonomous vehicles,'' \emph{Transportation Research Part C:
  Emerging Technologies}, vol.~68, pp. 126--143, 2016.

\bibitem{nv2020study}
R.~Nv-er, J.~Rong, and C.~Xu, ``Study on simulation of warning method and
  evaluation of car-following for intelligent drive,'' in \emph{IOP Conference
  Series: Materials Science and Engineering}, vol. 717.\hskip 1em plus 0.5em
  minus 0.4em\relax IOP Publishing, 2020, p. 012014.

\bibitem{yang2019efficient}
X.~Yang, D.~Zou, L.~Pei, D.~Sartori, and W.~Yu, ``An efficient simulation
  platform for testing and validating autonomous navigation algorithms for
  multi-rotor uavs based on unreal engine,'' in \emph{China Satellite
  Navigation Conference (CSNC) 2019 Proceedings: Volume II}.\hskip 1em plus
  0.5em minus 0.4em\relax Springer, 2019, pp. 527--539.

\bibitem{chae2018virtual}
B.~H. Chae, H.~K. Chae, and J.~Y. Lee, ``Virtual reality driving simulation for
  evaluation of road safety facilities,'' \emph{Journal of Digital
  Convergence}, vol.~16, no.~7, pp. 249--257, 2018.

\bibitem{liu2002algorithm}
J.~Liu and H.~Lan, ``An algorithm using predictive control theory for automatic
  drive,'' in \emph{Proceedings. International Conference on Machine Learning
  and Cybernetics}, vol.~3.\hskip 1em plus 0.5em minus 0.4em\relax IEEE, 2002,
  pp. 1601--1604.

\bibitem{bhattarai2020deep}
M.~Bhattarai and M.~Martinez-Ramon, ``A deep q-learning based path planning and
  navigation system for firefighting environments,'' \emph{arXiv preprint
  arXiv:2011.06450}, 2020.

\bibitem{rao2020research}
Y.~Rao and F.~Yang, ``Research on path tracking algorithm of autopilot vehicle
  based on image processing,'' \emph{International Journal of Pattern
  Recognition and Artificial Intelligence}, vol.~34, no.~05, p. 2054013, 2020.

\bibitem{leudet2019ailivesim}
J.~Leudet, F.~Christophe, T.~Mikkonen, and T.~M{\"a}nnist{\"o}, ``Ailivesim: An
  extensible virtual environment for training autonomous vehicles,'' in
  \emph{2019 IEEE 43rd annual computer software and applications conference
  (COMPSAC)}, vol.~1.\hskip 1em plus 0.5em minus 0.4em\relax IEEE, 2019, pp.
  479--488.

\bibitem{han2021new}
I.~Han, D.-H. Park, and K.-J. Kim, ``A new open-source off-road environment for
  benchmark generalization of autonomous driving,'' \emph{IEEE Access}, vol.~9,
  pp. 136\,071--136\,082, 2021.

\end{thebibliography}
\end{document}